\documentclass[times,final]{elsarticle}
\journal{Journal of Computational Physics}
\usepackage[T1]{fontenc}
\usepackage{amsmath,amssymb,mathtools}
\usepackage{graphicx}
\graphicspath{{figures/}}
\usepackage{booktabs,longtable,array,calc,tabularx}
\usepackage{microtype}
\usepackage{xurl}
\usepackage{hyperref}

\usepackage{geometry}
\makeatletter
\AtBeginDocument{%
  \fontsize{9.5pt}{11.4pt}\selectfont
}

\global\bibsep=0pt

\long\def\MaketitleBox{%
  \resetTitleCounters
  \def\baselinestretch{1}%
  \begin{flushleft}
    {\fontsize{16pt}{18.5pt}\selectfont\@title\par}
    \vskip12pt
    \ifdoubleblind
      \vspace*{2pc}%
    \else
      {\fontsize{10pt}{12pt}\selectfont\elsauthors\par}
      \vskip6pt
      {\fontsize{8.5pt}{10pt}\itshape\elsaddress\par}
      \vskip14pt
    \fi
  \end{flushleft}
  \noindent\rule{\textwidth}{0.35pt}\par\vskip9pt
  \noindent
  \begin{minipage}[t]{0.285\textwidth}
    \vspace{0pt}
    {\fontsize{8pt}{9.6pt}\selectfont\bfseries A R T I C L E\quad I N F O\par}
    \vskip5pt\hrule\vskip7pt
    {\fontsize{8pt}{10pt}\selectfont\textit{Keywords:}~\JCPKeywordText}
  \end{minipage}\hfill
  \begin{minipage}[t]{0.665\textwidth}
    \vspace{0pt}
    {\fontsize{8pt}{9.6pt}\selectfont\bfseries A B S T R A C T\par}
    \vskip5pt\hrule\vskip7pt
    {\fontsize{8.7pt}{10.7pt}\selectfont\JCPAbstractText}
  \end{minipage}
  \par\vskip10pt\noindent\rule{\textwidth}{0.35pt}\par\vskip11pt
}

\long\def\@makecaption#1#2{%
  \vskip\abovecaptionskip
  {\footnotesize\sbox\@tempboxa{#1.~#2}%
   \ifdim\wd\@tempboxa>\hsize
     #1.~#2\par
   \else
     \global\@minipagefalse
     \hb@xt@\hsize{\hfil\box\@tempboxa\hfil}%
   \fi}
  \vskip\belowcaptionskip}
\makeatother


\biboptions{sort&compress}
\begin{document}
\begin{frontmatter}
\long\gdef\JCPAbstractText{Neural operators provide fast learned surrogates for time-dependent partial differential equations (PDEs), but autoregressive deployment creates a refinement-allocation problem that is largely orthogonal to the choice of operator backbone: prediction errors vary over space and physical time, and only a finite number of local corrections may be committed to the trajectory. We formulate this setting as budgeted adaptive neural-operator solving, where the budget counts committed refinement interventions rather than raw floating-point operations. A global Fourier neural operator advances the full field, a local neural-operator corrector produces patch-wise residual proposals, and a conditional set-aware selector determines where corrections should be applied. The remaining decision - when and how much of the refinement-intervention budget to allocate - is treated as a sequential decision problem over physical time. We introduce rollout-verified policy improvement (RV-PI), which enumerates feasible refinement counts, evaluates their consequences through actual continuation rollouts of the learned PDE solver, converts verified long-horizon advantages into conservative soft policy targets, and accepts an update only when held-out trajectory error improves. All local proposals for a transition are generated once from the same provisional state and then frozen, ensuring that multi-patch actions have order-independent semantics. On the official shallow-water benchmark with a 32-intervention budget, RV-PI achieves a three-seed mean trajectory relative \(L_2\) error of 0.6910, improving over immediate-only policy improvement by 5.37\% and over RandomMacro by 2.41\%. On a stabilized forcing-driven Brusselator benchmark with a 76-intervention budget, RV-PI attains the best three-seed mean trajectory error, 0.09954, improving over immediate-only policy improvement by 2.31\% and over RandomMacro by 5.32\%. These results support a solver-level principle: under a fixed refinement-intervention budget, the value of a local correction depends on its downstream effect on the autoregressive trajectory, not only on its immediate error reduction.}
\long\gdef\JCPKeywordText{neural operator; partial differential equations; adaptive refinement; reinforcement learning; Fourier neural operator; long-horizon rollout; local correction}
\title{Learning When to Refine: Long-Horizon Reinforcement Learning for Budgeted Neural-Operator PDE Solvers}
\author{Ange Tong}
\address{National Research Tomsk State University, 36 Lenin Avenue, Tomsk 634050, Russian Federation}

\end{frontmatter}

\section{Introduction}\label{introduction}

Neural operators have changed the computational role of machine learning in PDE simulation. Rather than fitting a separate finite-dimensional map for each discretization, operator-learning models approximate mappings between function spaces and can be reused across families of inputs. Fourier neural operators (FNOs), DeepONets, and related operator-learning architectures have demonstrated that expensive numerical solution maps can be approximated with low online cost once training is complete \cite{ref1,ref2,ref3}. This capability is attractive in repeated-query settings such as uncertainty propagation, inverse design, digital twins, and real-time forecasting. Yet a learned time-marching solver introduces a different difficulty from a one-shot surrogate: its own prediction becomes part of the next input, so local approximation errors can alter future states and accumulate over the rollout.

\pagebreak

For time-dependent problems, however, the operator is only one component of the deployed algorithm. A practical learned solver repeatedly composes the approximation with itself, so its input distribution is partly generated by its own past errors. This distinction is central in the literature on autoregressive neural PDE solvers: a model can be accurate under teacher-forced one-step evaluation and yet drift when repeatedly exposed to self-generated states \cite{ref18,ref19,ref26}. The numerical-analysis analogue is familiar - local truncation quality does not by itself determine the behavior of a long integration - but learned time steppers add a distribution-shift mechanism because the approximation changes the states on which the next approximation is evaluated. Consequently, solver design must address both representation error and how that error is managed over the physical trajectory.

Recent work has attacked this long-horizon problem from several directions. PDE-Refiner improves spectral fidelity through iterative denoising-style refinement \cite{ref18}; message-passing neural PDE solvers and unrolled training expose the model to its own predictions to improve stability \cite{ref19,ref20}; hybrid neural-numerical approaches use learned operators to warm start, correct, or accelerate conventional solvers \cite{ref12,ref22}; and time-adaptive operator learning changes the temporal discretization itself \cite{ref23}. These approaches motivate a complementary question that becomes important once a deployable global predictor and a local correction mechanism already exist: how should a finite number of corrections be distributed over the trajectory? Our work isolates this allocation problem rather than proposing another operator backbone.

The resulting question is not simply whether a neural operator can predict the next state. It is whether a learned solver can decide how to allocate a limited number of corrective interventions over an evolving trajectory. Suppose a global operator provides a full-field transition and a local operator can propose residual corrections on spatial patches. Committing every available correction at every transition defeats the purpose of a budgeted solver, whereas a fixed regular schedule ignores the fact that error sensitivity changes with the evolving state. The natural objective is therefore constrained: minimize error accumulated over the complete physical trajectory subject to a finite budget of applied local corrections. In the present implementation, the proposal pool is materialized in a batched local-operator pass so that all candidate patches share the same provisional state; accordingly, the budget should be interpreted as a refinement-intervention budget rather than as a direct measurement of end-to-end FLOPs. This distinction is kept explicit throughout the paper.

This objective separates into two coupled decisions. The spatial decision asks where local correction is valuable at the current transition. The temporal decision asks when and how much of the remaining budget should be spent. The first can be estimated from state-dependent patch features and correction proposals. The second is intrinsically sequential: a correction at time t changes the state presented to the global operator at t+1, thereby changing the entire downstream trajectory. Immediate error reduction is therefore not generally equivalent to long-horizon value. This is the gap addressed in this work.

The decomposition also separates two sources of generalization difficulty. Spatial selection must distinguish patches whose frozen residual proposals are useful given the corrections already selected, which makes independent top-k scoring insufficient when patches overlap or address redundant structures. Temporal allocation must distinguish transitions whose corrections have similar immediate effect but different future consequences. The latter is the point at which reinforcement learning becomes scientifically motivated: the action modifies the solver state, and that modified state changes the future transition distribution. The environment is therefore not an external control benchmark attached to a PDE model; it is the learned PDE solver itself.

We develop a hierarchical fixed-grid solver that explicitly decomposes these decisions. A standard conditional FNO predicts a provisional next state. A local FNO then generates one residual-correction proposal for every candidate patch from that same provisional state. These proposals are frozen. A transformer-based set-aware selector chooses a patch set conditionally, so the score of a candidate can depend on patches already selected. A macro policy chooses \(q_t\in\{0,1,2,4\}\), the number of local corrections to spend at transition t, under a hard trajectory budget. The selected frozen corrections are blended simultaneously through normalized overlap weights, producing the corrected state that is fed to the next physical transition.

The main algorithmic contribution is rollout-verified policy improvement (RV-PI). The macro policy is initialized from a train-only beam-search teacher, but it is not restricted to teacher imitation. At states sampled from several deployable policies, RV-PI enumerates every feasible refinement count and executes the actual learned PDE solver forward to the end of the trajectory under the current policy. These verified continuation returns define action advantages. The policy is updated by an advantage-weighted target while retaining behavioral and Kullback-Leibler regularization to the previous policy. The candidate update is then evaluated on a held-out validation split and rolled back unless trajectory relative \(L_2\) error decreases. An immediate-only version differs only by truncating the return after the current transition, giving a controlled test of whether future consequences matter.

RV-PI is deliberately closer to verified approximate policy iteration than to generic model-free reinforcement learning. Candidate action values are measured by replaying the actual global operator, frozen local proposals, set-aware selector, and current continuation policy through the remaining physical horizon. This direct verification aligns the policy target with the trajectory-level objective and makes temporal credit assignment auditable: the source of an update is an explicit solver-executed continuation return, and the updated policy is checked again on a held-out validation split.

Two benchmark families stress different aspects of the formulation. The shallow-water experiment uses the official condition field and 72-frame, \(256\times 256\) trajectories distributed with the Laplace Neural Operator benchmark. The forcing-driven Brusselator experiment uses 39-frame, \(28\times 28\) trajectories with a time-varying scalar forcing. Brusselator additionally motivates deployment-aligned operator training: acceptable one-step accuracy alone is insufficient for reliable recursive rollout. A six-step closed-loop objective with a mild amplitude guard changes validation trajectory relative \(L_2\) from 8.2505 to 0.09369 while also improving one-step error, establishing a stable autoregressive backbone for the refinement-allocation study.

Together, the two benchmarks provide a controlled transfer test across substantially different rollout structures, spatial resolutions, context representations, patch geometries, and baseline error scales. The shallow-water experiment operates on a high-resolution benchmark with a static released condition field, whereas the Brusselator release is a compact forcing-driven reaction-diffusion trajectory with a time-varying scalar input. The policy architecture and action set remain structurally the same, while the operator backbones, patch partitions, and total budgets follow the benchmark geometry.

The evidence is organized around the claim that motivates the method. Under matched local-call budgets, RV-PI attains the lowest three-seed mean trajectory error on both benchmark families. Relative to ImmediateOnlyPI, the reduction is 5.37\% on shallow water and 2.31\% on Brusselator. Because the operator backbone, spatial selector, action set, and total applied-refinement budget are controlled, this comparison isolates the value of full-horizon temporal allocation. The contribution is therefore a solver-level formulation and learning procedure for allocating finite learned refinement over physical time.

\textbf{The main contributions are:}

\begin{itemize}
\item
  A finite-budget formulation of adaptive neural-operator time marching in which spatial refinement and temporal refinement allocation are separated while remaining coupled through the evolving solver state.
\item
  A hierarchical fixed-grid solver that provides frozen local correction proposals, conditional set-aware patch selection, and simultaneous normalized overlap blending as a well-defined deployment substrate for temporal allocation.
\item
  Rollout-verified policy improvement, a conservative long-horizon policy-update procedure that evaluates feasible refinement counts through true future solver rollouts and yields lower mean trajectory error than immediate-only allocation on two distinct PDE benchmark families under identical local-call budgets.
\end{itemize}

\begin{figure}[!t]
\centering
\includegraphics[width=\textwidth]{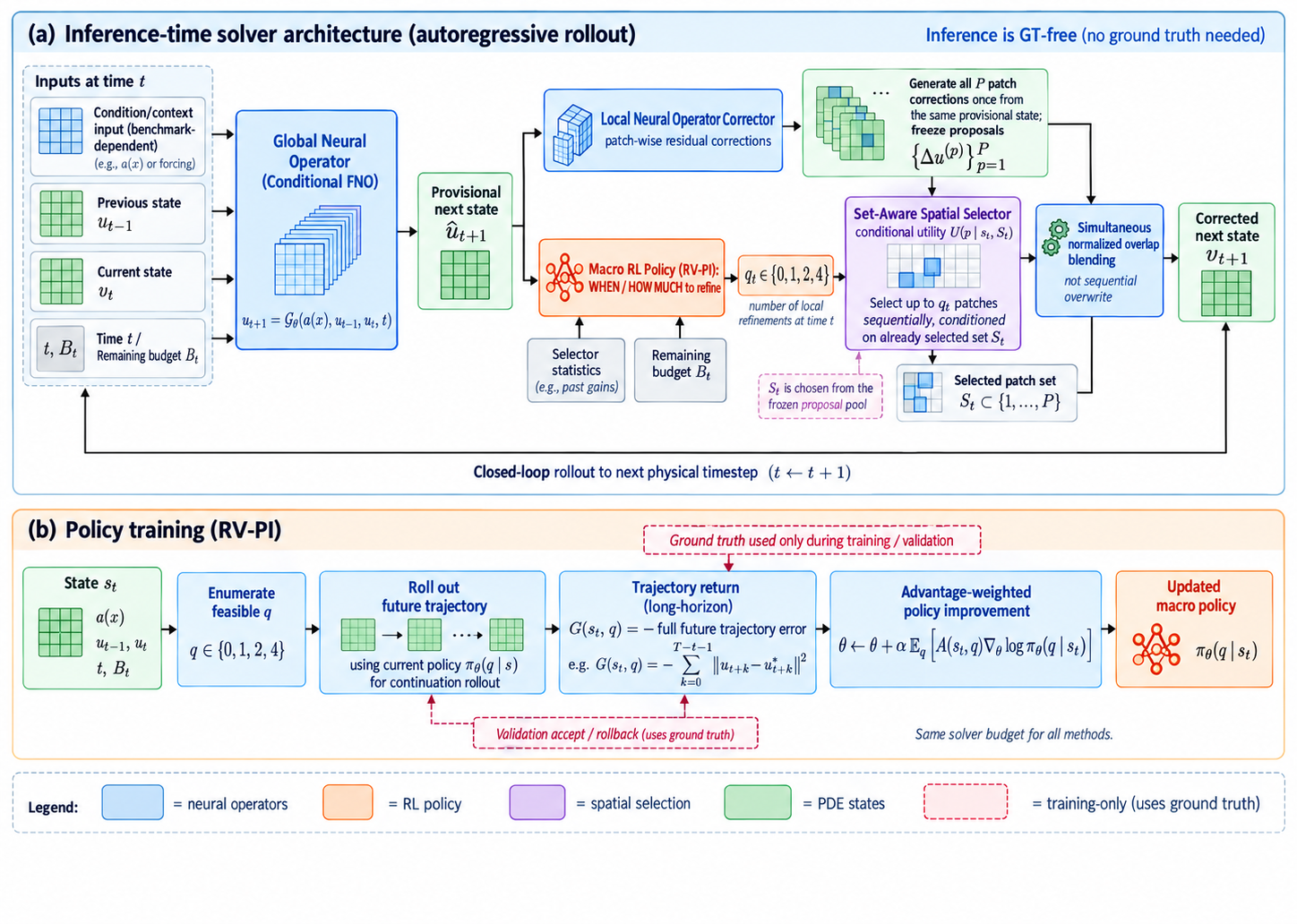}
\caption{Overview of the proposed budgeted adaptive neural-operator solver. The global operator produces a provisional transition; all local patch corrections are generated once from that state and frozen. The macro RV-PI policy decides WHEN/HOW MUCH to refine, while the conditional selector decides WHERE. Selected frozen corrections are applied simultaneously by normalized overlap blending. The trajectory budget counts committed correction interventions; ground truth is used only to construct training-time selector gains, rollout returns, and validation acceptance, and inference is ground-truth free. The initial schematic layout was prepared with assistance from OpenAI image-generation tools and subsequently edited and verified by the author against the implemented solver.}
\label{fig:architecture}
\end{figure}

\section{Related work and positioning}\label{related-work-and-positioning}

\subsection{Neural operators and autoregressive learned solvers}\label{neural-operators-and-autoregressive-learned-solvers}

Operator learning seeks reusable approximations to maps between function spaces. DeepONet represents nonlinear operators through coupled branch and trunk networks, while the neural-operator framework formalizes discretization-invariant operator approximation and includes Fourier parameterizations \cite{ref1,ref2,ref3}. The Fourier neural operator is particularly effective on regular grids because global convolution can be performed efficiently in spectral space. Extensions such as Geo-FNO address geometric flexibility \cite{ref5}, and physics-informed neural operators incorporate governing-equation constraints \cite{ref6}. Laplace neural operators target transient and non-periodic responses and provide the official shallow-water and Brusselator datasets used here \cite{ref4}. These works primarily improve the operator representation or its training objective. In contrast, the present study intentionally uses compact FNO backbones and asks how an already trained learned solver should allocate additional corrective interventions during deployment.

For the present problem, two properties of operator learning matter more than architectural novelty. First, an operator model can be reused over a family of initial conditions, forcings, or coefficient fields, which makes offline investment in training and policy optimization meaningful when the solver is queried repeatedly. Second, the operator can be embedded inside a larger numerical procedure rather than treated as a monolithic replacement for simulation. Recent JCP work on locality explicitly shows that physical locality can be enforced through decomposition rather than left to a global network to discover implicitly \cite{ref21}. The Local Neural Operator similarly emphasizes reusable localized evolution rules on varied domains \cite{ref13}. These observations support our decision to treat the global and local operators as solver components with distinct roles.

We instantiate both the global and local predictors with compact FNOs because the Fourier architecture is established, efficient, and easy to condition on state history and exogenous fields \cite{ref1,ref2}. Holding the operator family and capacity fixed makes the central comparison diagnostic: changes in accuracy can be attributed to refinement allocation rather than to a larger or more specialized backbone. The proposed framework is consequently compatible with other neural-operator families so long as they expose a global transition and a local correction mechanism.

\subsection{Local and iterative correction of learned PDE predictions}\label{local-and-iterative-correction-of-learned-pde-predictions}

A growing body of work augments global learned predictions with iterative or local correction. Local neural operators exploit locality to construct reusable time-marching components on varied domains \cite{ref13}, while hybrid neural/numerical solvers use learned operators to complement classical relaxation \cite{ref12}. More recently, iterative-refinement neural operators have treated post-prediction correction as a learned fixed-point process \cite{ref11}. These approaches establish that correction can be a first-class component of a learned solver. Our setting differs in the resource constraint: local correction is available, but not at every location and time. The central problem is therefore selection and scheduling under a finite budget rather than repeated full-field refinement.

Iterative-refinement methods address a related but different axis. PDE-Refiner repeatedly denoises a prediction so that low-amplitude spectral components are not neglected during long rollouts \cite{ref18}. IRNO interprets learned refinement as a shared-weight fixed-point iteration and provides a contraction-based view of convergence \cite{ref11}. Blended neural-operator/relaxation methods and super-fidelity strategies instead couple a learned approximation to an iterative numerical process \cite{ref12,ref22}. In each case, additional corrective work is used to improve fidelity. Our question begins one level above that mechanism: given that a local correction operator is already available, which physical transitions should receive a limited number of corrections, and which spatial corrections should be committed at those transitions?

This difference also explains why the frozen-proposal construction is more than an implementation convenience. Iterative refinement typically allows the next correction to depend on the previously corrected state. That is appropriate when the iteration count itself defines the algorithm. In our setting, however, the macro action \(q_t\) is supposed to mean that \(q_t\) comparable corrections are committed at transition t. Regenerating candidates after each patch would change the meaning and cost of later choices. We therefore generate all candidates from one provisional state, freeze them, and let the selector reason over a fixed candidate set.

\subsection{Adaptive mesh refinement and reinforcement learning}\label{adaptive-mesh-refinement-and-reinforcement-learning}

Adaptive mesh refinement (AMR) allocates numerical resolution where it is needed. Reinforcement-learning formulations of AMR have highlighted exactly the long-horizon structure that heuristic error indicators can miss: refinement decisions alter the future simulation, so immediate error estimates need not optimize whole-trajectory performance \cite{ref8,ref9}. Deep RL has also been used to learn transferable AMR policies across PDEs and discretizations. In 2026, RLMesh further used reinforcement learning to allocate mesh points during surrogate-data generation \cite{ref10}. Our work is complementary but technically distinct. We keep the physical grid fixed, do not remesh or alter the discretization, and allocate calls to a learned local correction operator during neural-solver inference. The action is a refinement count over time, while a separate learned selector resolves spatial location.

Several recent methods broaden adaptive computation beyond classical mesh error indicators. RL-guided mesh optimization learns where to place simulation or surrogate-training resolution \cite{ref10}, while time-adaptive operator learning changes the physical step size to follow temporal complexity \cite{ref23}. These approaches and ours share a resource-allocation perspective but act on different computational objects. Mesh methods change discretization; TANTE changes temporal step size; our solver keeps the released grid and transition cadence fixed and allocates which learned local corrections are committed. The distinction matters for deployment scenarios in which the discretization is fixed by an existing data product, sensor grid, or surrogate interface.

The most directly relevant RL-AMR studies show why myopic error reduction can be insufficient. Refining one region changes the future numerical trajectory and therefore changes which later refinements become useful \cite{ref8,ref9}. Our method retains this sequential perspective but moves it into a learned fixed-grid solver. The state contains learned-field context, selector statistics, and remaining budget rather than hand-engineered mesh indicators, and the action selects the number of local corrections rather than mesh cells or refinement levels.

\subsection{Advantage-weighted and conservative policy improvement}\label{advantage-weighted-and-conservative-policy-improvement}

The RV-PI update is related in spirit to advantage-weighted regression and conservative policy-improvement ideas \cite{ref14,ref15,ref24,ref25}: actions with larger measured advantage receive larger supervised weight, while behavioral and Kullback-Leibler regularization keep each update close to the policy that generated the state distribution. RV-PI replaces critic-based value estimation with explicit continuation rollouts of the learned PDE solver. The validation accept/rollback step then adds direct trajectory-level selection on top of the conservative update, yielding a policy-improvement procedure whose targets and acceptance criterion are both expressed in the solver's deployed objective.

\subsection{Long-horizon stability and temporal adaptation}\label{long-horizon-stability-and-temporal-adaptation}

Autoregressive stability is a prerequisite for meaningful allocation. Classical one-step training observes ground-truth states, whereas inference exposes the model to states generated by its own approximation. Geneva and Zabaras \cite{ref26} showed early that autoregressive PDE surrogates can accumulate uncertainty and error over repeated application. Brandstetter et al. \cite{ref19} cast stability partly as a distribution-shift problem, and PDE-Refiner \cite{ref18} demonstrated that seemingly small spectral inaccuracies can become decisive over long horizons. More recent systematic studies of unrolled training show that exposing a model to self-generated trajectories can substantially improve inference behavior even when gradients are truncated through parts of the rollout \cite{ref20}.

This literature directly motivates the Brusselator V2 backbone. A one-step-oriented model achieves moderate local accuracy yet develops large recursive drift over a 38-step rollout. Aligning coarse-model training with deployment through a six-step closed-loop objective restores stable autoregressive behavior while leaving the policy design unchanged. The resulting separation is methodologically useful: operator training establishes a viable closed-loop transition model, after which the policy experiment isolates the temporal allocation problem.

Temporal adaptation can also be implemented by modifying the solver clock. TANTE \cite{ref23}, for example, learns adaptive time stepping through a continuous-time operator construction. Our method leaves the benchmark time grid unchanged and adapts the density of local corrections instead. The two strategies are complementary: adaptive time stepping changes when the global state is advanced, whereas RV-PI changes when additional local corrective capacity is committed. This distinction helps position the present contribution within the broader trend toward adaptive scientific machine learning.

\section{Problem formulation}\label{problem-formulation}

\subsection{Autoregressive operator time marching}\label{autoregressive-operator-time-marching}

Let \(u_t\) denote the resolved PDE state at discrete physical transition t and let \(c_t\) denote benchmark-dependent exogenous context. For shallow water, \(c_t\) is the official static condition field \(a(x,y)\). For Brusselator, \(c_t\) is the known scalar forcing at transition t. A global learned transition operator \(G_\theta\) produces a provisional next state from recent state history and time context:

\begin{equation}
\hat{u}_{t+1}=G_{\theta}(c_t,u_{t-1},u_t,u_t-u_{t-1},\tau_t).
\end{equation}

The implementation uses a compact conditional FNO. The provisional state is not yet the final solver output; it is the common reference from which all local candidates are generated.

\subsection{Frozen local refinement proposals}\label{frozen-local-refinement-proposals}

Let \(\mathcal{P}=\{1,\ldots,P\}\) index a fixed patch partition. For every patch \(p\), a local operator \(L_\phi\) predicts a residual correction from the same provisional field and local state/context features:

\begin{equation}
\Delta u_t^{(p)}=L_{\phi}(c_t,u_{t-1},u_t,\hat{u}_{t+1},\hat{u}_{t+1}-u_t,u_t-u_{t-1},\tau_t;p).
\end{equation}

All P proposals are computed once and frozen before any patch is selected. If \(S_t\) is the selected set, the corrected transition is obtained by simultaneous normalized overlap blending, denoted \(\mathcal{B}\):

\begin{equation}
u_{t+1}=\mathcal{B}\!\left(\hat{u}_{t+1},\{\Delta u_t^{(p)}:p\in S_t\}\right).
\end{equation}

The frozen-proposal rule is important because it makes the cost and semantics of a local call independent of selection order. Choosing a second patch does not regenerate the first proposal from an already corrected state. Consequently, the macro action \(q_t\) counts comparable refinement units.

\begin{equation}
u_{t+1}(x)=\hat{u}_{t+1}(x)+\frac{\sum_{p\in S_t}w_p(x)\Delta u_t^{(p)}(x)}{\max\!\left(\sum_{p\in S_t}w_p(x),\varepsilon\right)}.
\end{equation}

Here \(w_p\) is a deterministic tapering window associated with the patch halo. The implementation uses a separable sine ramp in the halo and unit weight in the patch core. Normalization by the accumulated weight makes overlap blending simultaneous: the result depends on the selected set but not on an arbitrary write-back order. This is the state that enters the next physical transition.

\subsection{Conditional set selection}\label{conditional-set-selection}

The spatial decision is formulated as conditional set construction. Let \(S\) denote the patches already selected at the current transition. For any unselected candidate \(p\), define the training-time marginal gain

\begin{equation}
g_t(p\mid S)=\ell\!\left(\mathcal{B}(\hat{u}_{t+1},S),u^\ast_{t+1}\right)-\ell\!\left(\mathcal{B}(\hat{u}_{t+1},S\cup\{p\}),u^\ast_{t+1}\right),
\end{equation}

where $\ell$ is the field mean-squared error used to construct selector labels and \(\mathcal{B}\) denotes the same frozen-proposal blending operator used at inference. Positive \(g_t\) means that adding \(p\) reduces immediate field error conditioned on what is already selected. The selector learns a score $U_{\psi}(p\mid s_t,S)$ from deployable field features, proposal statistics, a selected-set mask, and compact context. At inference, the selected set is built greedily by repeatedly choosing the highest-scoring unselected patch until \(|S_t|=q_t\). Ground truth is used to construct \(g_t\) on the training split, but neither \(g_t\) nor \(u^\ast_{t+1}\) is available to the deployed selector.

\subsection{Finite-budget sequential decision problem}\label{finite-budget-sequential-decision-problem}

At each transition the macro controller selects the number of local corrections

\begin{equation}
q_t\in\mathcal{A}=\{0,1,2,4\},\qquad |S_t|=q_t.
\end{equation}

A hard budget B is imposed over the trajectory. The implementation masks actions that would make the remaining budget impossible to exhaust using the remaining transitions, so compared adaptive methods use exactly the same number of local calls:

\begin{equation}
\sum_{t=0}^{T-1}q_t=B.
\end{equation}

The optimization target is trajectory-integrated prediction error, represented here by a normalized squared-error objective equivalent to the return used during policy improvement:

\begin{equation}
J(\pi)=\mathbb{E}_{\pi}\!\left[\frac{\sum_t\|u_{t+1}-u^\ast_{t+1}\|_2^2}{\sum_t\|u^\ast_{t+1}\|_2^2}\right],\qquad \text{subject to }\sum_t q_t=B.
\end{equation}

This formulation makes the temporal decision long-horizon by construction. The value of \(q_t\) depends not only on its immediate correction but also on how the corrected \(u_{t+1}\) changes all subsequent global and local transitions.

The hard equality budget is enforced through feasibility masking rather than a reward penalty. If \(B_t\) is the remaining budget and \(H_t\) is the number of transitions remaining after the current decision, define \(\mathcal{R}(H_t)\) as the set of totals reachable by summing \(H_t\) actions from \(\mathcal{A}\). The feasible action set is

\begin{equation}
\mathcal{A}_t(B_t)=\{q\in\mathcal{A}:0\le q\le B_t\ \text{and}\ (B_t-q)\in\mathcal{R}(H_t)\}.
\end{equation}

This construction guarantees that a trajectory which starts with budget B can always finish with exactly that budget exhausted. It prevents a policy from obtaining an apparent accuracy advantage by simply using fewer or more committed corrections than a baseline. It also removes the need to tune a Lagrange multiplier that would trade accuracy against budget violation.

\subsection{Deployment contract and information boundary}\label{deployment-contract-and-information-boundary}

The solver is designed around a strict separation between training-only supervision and deployable inputs. During training, ground truth is used to supervise the global and local operators, to construct conditional marginal-gain labels for the selector, to score train-only beam-search trajectories, and to compute verified continuation returns for policy improvement. During validation, ground truth is used only to decide whether a candidate policy update reduces trajectory error. At test-time inference, the deployed computation consists only of the learned global operator, the local proposal operator, the set-aware selector, the macro policy, and bookkeeping for time and remaining budget. No ground-truth field, residual to the unknown solution, or oracle patch label is an inference input.

A second deployment contract concerns the meaning of the refinement budget. The implementation materializes every local proposal from the common provisional state in one batched pass because the selector needs a frozen candidate pool. The quantity B therefore counts corrections committed to the trajectory - the number of patches whose proposed residuals are actually blended into states - rather than measuring total floating-point operations. This abstraction isolates the sequential allocation question while preserving exact fairness in committed refinements. We do not claim that the present prototype reduces wall-clock time in proportion to B; a lazy proposal implementation would be required to convert the same policy into a strict FLOP or latency budget.

For computational-complexity accounting, let $C_G$ denote one global-operator transition, $C_L$ one local-patch proposal, $C_S$ one selector evaluation over the frozen candidate pool, and $C_\pi$ one macro-policy evaluation. With $P$ candidate patches and action $q_t$, the current prototype performs work of order $C_G + P C_L + q_t C_S + C_\pi$ per transition, with the $P$ local proposals executed as a batch. Over a $T$-transition rollout, the proposal term therefore scales with $TP$ independently of the committed budget, whereas $\sum_t q_t=B$ controls the number of selector decisions whose proposals are actually blended into the propagated state. This accounting makes explicit why the present experiments evaluate allocation quality under a matched intervention budget rather than claim a proportional wall-clock speedup.

\section{Budgeted adaptive neural-operator solver}\label{budgeted-adaptive-neural-operator-solver}

\subsection{Global and local Fourier neural operators}\label{global-and-local-fourier-neural-operators}

The global and local predictors are intentionally standard FNOs rather than newly proposed operator backbones. This choice isolates the contribution of adaptive computation from architectural capacity. On shallow water, the global FNO receives the official condition field, previous state, current state, temporal difference, and normalized time. It predicts at \(64\times 64\) and is bilinearly lifted to the native \(256\times 256\) grid for refinement. The local FNO receives patch crops of the condition, state history, provisional state, two temporal differences, and time, and directly predicts a residual correction. On Brusselator, the corresponding exogenous input is the scalar time-varying forcing, and both global and local models operate on the \(28\times 28\) field.

For shallow water, the global model input concatenates the normalized official condition field, the previous state, the current state, the temporal difference \(u_t-u_{t-1}\), and a scalar time channel. The conditional FNO therefore receives \(\texttt{condition\_channels}+3\,\texttt{state\_channels}+1\) input channels. The global prediction is produced on a \(64\times 64\) working grid and bilinearly returned to the native \(256\times 256\) field for refinement. The local model is conditioned on the same history together with the provisional next state and both temporal differences, so its residual target is explicitly anchored to the transition proposed by the global operator. This organization makes the local network a correction model rather than an independent next-state predictor.

The shallow-water local targets are residuals $u^\ast_{t+1}-\hat{u}_{t+1}$. Training states deliberately mix teacher-forced, coarse closed-loop, and randomly refined closed-loop trajectories in proportions 0.50/0.25/0.25. This mixture exposes the local corrector to the kinds of perturbed states that it will encounter when the macro policy changes the refinement schedule. The Brusselator local corrector uses the same principle: half of sampled states are teacher-forced and the remaining half are divided between coarse and locally corrected closed-loop states. Thus the corrector is trained to operate around the deployed state distribution rather than only around exact trajectories.

The shallow-water configuration uses width 32, 12 retained Fourier modes, and depth 4 for both coarse and local operators. The native field is divided into an \(8\times 8\) patch grid with \(32\times 32\) cores and an 8-cell halo. Brusselator V2 uses a width-32, 10-mode, depth-4 global FNO and a width-32, 8-mode, depth-3 local FNO, with a \(7\times 7\) patch grid, \(4\times 4\) cores, and a one-cell halo. The models are therefore small enough that the study focuses on scheduling rather than model scaling.

\subsection{Closed-loop training of the operator backbone}\label{closed-loop-training-of-the-operator-backbone}

A budget policy is meaningful only if the uncontrolled learned transition is itself a viable closed-loop dynamical model. The shallow-water coarse operator is trained with one-step supervision plus a two-step term and progressively exposes its second transition to its own predicted state. The local corrector is trained on a mixture of teacher-forced, coarse-rollout, and locally refined states. For Brusselator V2, the instability of a one-step-trained baseline is more pronounced, so the coarse loss is explicitly multi-step. Six-step autoregressive windows are optimized together with a one-step anchor and a mild penalty when predicted root-mean-square amplitude exceeds a multiple of the target amplitude. Model selection uses trajectory error on a validation subset. This training is not part of the RV-PI contribution; it is a solver-quality gate that ensures the adaptive controller is evaluated on a meaningful environment.

The stabilized Brusselator objective is fully autoregressive over a short training horizon K=6. Starting from a ground-truth anchor pair, the model recursively feeds its own prediction into subsequent transition inputs. The loss combines the mean multi-step rollout error, a one-step anchor, and a weak amplitude-ratio penalty:

\begin{equation}
\begin{aligned}
\mathcal{L}_{\mathrm{coarse}}={}&\frac{1}{K_{\mathrm{eff}}}\sum_{k=1}^{K_{\mathrm{eff}}}\mathrm{MSE}(\hat{u}_{t+k},u^\ast_{t+k})+0.5\mathcal{L}_1\\
&+0.01\frac{1}{K_{\mathrm{eff}}}\sum_k\left[\max\!\left(\frac{\mathrm{RMS}(\hat{u}_{t+k})}{\mathrm{RMS}(u^\ast_{t+k})}-2,0\right)\right]^2.
\end{aligned}
\end{equation}

The one-step anchor prevents the model from obtaining apparent long-horizon stability by collapsing to an over-smoothed predictor. The amplitude term is only an explosion guard: it is inactive while prediction RMS remains within twice the target RMS. Training runs for 16 epochs with batch size 8 and AdamW at \(5\times 10^{-4}\). Every two epochs the model is rolled out on a fixed 32-case validation subset, and the checkpoint with the lowest validation trajectory relative \(L_2\) is retained. Both the diagnostic V1 and the V2 evaluation use the same finite-value environment guard, so the large stability improvement reported later is not attributable to a new clipping rule.

Shallow water does not require the same stabilization intervention. Its coarse FNO is trained for 12 epochs with one-step supervision plus a two-step term whose second input is increasingly replaced by the model prediction during the latter half of training. The contrast between the two benchmarks is useful: the allocation framework does not mandate one universal operator-training recipe; it requires only that the backbone passed to policy learning behave as a viable closed-loop transition model.

\subsection{Set-aware spatial selection: learning WHERE to refine}\label{set-aware-spatial-selection-learning-where-to-refine}

Given a frozen proposal pool, the spatial selector estimates the conditional marginal value of each unselected patch. Its input combines downsampled solver fields, patch-level statistics derived from the predicted corrections and provisional field, geometric/overlap features, an embedding indicating which patches have already been selected, and solver context. A transformer encoder exchanges information among patch tokens. The resulting score for candidate p is therefore conditioned on the current set S:

Architecturally, the selector has two information streams. A small convolutional field encoder compresses solver fields, while an MLP maps per-patch statistics into a 64-dimensional token space. A binary selected-mask embedding marks whether a candidate is already part of S, and a three-dimensional context vector is projected into the same token space. Two pre-norm Transformer encoder layers with four attention heads allow each patch token to interact with all other candidates before a scalar score is produced. This is where set awareness enters: the score of patch p can change after another patch is selected even though the underlying local proposals remain frozen.

The selector supervision combines three signals. A Huber regression term fits normalized marginal gains; a pairwise logistic ranking term preserves the ordering of candidates with different gains; and a margin term pushes the best valid candidate above the remaining alternatives. The purpose is not to predict the exact magnitude of the oracle gain with high calibration accuracy. It is to recover enough of the conditional ordering that a greedy set construction can identify complementary local corrections without access to ground truth at deployment.

\begin{equation}
U_{\psi}(p\mid s_t,S)\approx \text{immediate marginal error reduction of adding }p\text{ to }S.
\end{equation}

Training labels are constructed only on the training split by comparing the ground-truth mean-squared error before and after adding each frozen candidate to a partially selected set. The selector loss combines robust regression to normalized gains, pairwise ranking, and a margin term that separates the best valid patch. At inference, selection is sequential: at each of \(q_t\) substeps the network is reevaluated with the updated selected-set mask, and the highest-scoring remaining patch is added. Ground truth is never an inference input.

\subsection{Macro policy: learning WHEN and HOW MUCH to refine}\label{macro-policy-learning-when-and-how-much-to-refine}

The macro policy receives four groups of features: (i) solver fields including condition/forcing, previous state, current state, provisional state, and temporal differences; (ii) summary statistics of the selector score distribution, including top-1/top-2/top-4 means, standard deviation, and fraction of positive scores; (iii) the selector's pooled embedding; and (iv) normalized time, remaining budget, and remaining-horizon context. A small convolutional encoder and multilayer perceptron output logits over the four actions \(\{0,1,2,4\}\). Feasibility masking enforces the budget constraint exactly.

In code, the field tensor is encoded by two strided convolutions (24 and 48 channels) followed by global average pooling. The non-image part of the observation contains five selector-score summaries, the pooled 64-dimensional selector embedding, and three context scalars. The resulting vector is passed through hidden layers of sizes 128 and 64 to produce four action logits. This network is intentionally small relative to the neural operators. Its purpose is not to re-solve the PDE in latent space but to summarize whether the current state, spatial opportunity distribution, remaining budget, and remaining horizon justify spending 0, 1, 2, or 4 corrections now.

Feasibility masking is applied before every macro action, including during teacher search, RV-PI rollouts, and evaluation. The masking rule is stronger than the condition \(q_t\le B_t\): it also checks that the residual budget can be exactly represented by the remaining number of actions. Consequently, policy learning never sees an action that would make later exact budget satisfaction impossible. This property is important for fair comparisons because every adaptive trajectory terminates with the same number of committed corrections.

\subsection{Beam-search initialization}\label{beam-search-initialization}

Before policy improvement, a train-only beam search provides a stable initialization. The beam expands feasible macro actions while the spatial action is always generated by the deployable set-aware selector. Complete partial trajectories are ranked using training-set ground-truth squared error, with beam width 16. The best macro sequence forms behavior-cloning targets for BeamBC. This teacher is not used at test time and does not supply spatial ground-truth choices; it only gives an initial temporal schedule from which RV-PI can improve.

The teacher beam has width 16. For each partial trajectory, feasible q values are expanded, the set-aware selector supplies the corresponding spatial patches, and cumulative training-set squared error is used to rank the resulting partial trajectories. Only the macro action sequence is distilled; the selector is never replaced by an oracle spatial action. The behavior-cloned BeamBC policy is trained for 30 epochs and becomes the common initialization for both RV-PI and ImmediateOnlyPI. This shared initialization makes the horizon ablation substantially cleaner than comparing unrelated policies trained from scratch.

\subsection{Computational semantics and implementation invariants}\label{computational-semantics-and-implementation-invariants}

Three invariants are enforced throughout the implementation. First, frozen-proposal semantics ensure that all candidate patch residuals at a transition are functions of the same provisional state. Second, simultaneous normalized blending ensures that an action is a set, not an ordered list of overwrites. Third, exact budget masking ensures that methods compared at a fixed budget commit exactly the same number of corrections. These invariants turn the macro action into a stable experimental object: \(q_t\) has the same meaning independent of patch order, remaining-horizon accidents, or hidden regeneration of candidates.

The resulting per-transition structure consists of one global operator evaluation, one batched proposal evaluation over all patches, \(q_t\) sequential selector scoring passes, one compact macro-policy evaluation, and one normalized blend. The experiments therefore treat B as an exact intervention budget: every adaptive method commits the same number of patch corrections, while RV-PI changes only their temporal placement. This controlled abstraction isolates allocation quality from operator capacity and from the number of applied refinements; proposal-generation optimizations such as lazy evaluation or caching are orthogonal implementation choices.

\subsection*{Algorithm 1. Budgeted adaptive neural-operator rollout}
\noindent\textbf{Input:} trained global operator \(G_\theta\), local operator \(L_\phi\), selector \(U_\psi\), macro policy \(\pi\), budget \(B\).\\
\textbf{Initialize:} \(\texttt{previous}=\texttt{current}=u_0\), \(\texttt{remaining}=B\).\\
\textbf{For each physical transition \(t\):}
\begin{enumerate}
\item Compute provisional next state \(\hat{u}_{t+1}=G_\theta(\ldots)\).
\item Generate all patch corrections \(\{\Delta u_t^{(p)}\}\) from \(\hat{u}_{t+1}\); freeze the proposal pool.
\item Build macro observation; mask actions that violate exact remaining-budget feasibility.
\item Select \(q_t\in\{0,1,2,4\}\) from \(\pi\).
\item Starting from \(S=\varnothing\), select \(q_t\) patches sequentially using \(U_\psi(p\mid s_t,S)\).
\item Apply selected frozen corrections simultaneously with normalized overlap blending.
\item Set \(\texttt{previous}\leftarrow\texttt{current}\), then \(\texttt{current}\leftarrow u_{t+1}\), and \(\texttt{remaining}\leftarrow\texttt{remaining}-q_t\).
\end{enumerate}
\noindent\textbf{Return:} the corrected autoregressive trajectory.

\section{Rollout-verified policy improvement}\label{rollout-verified-policy-improvement}

\subsection{Verified continuation returns}\label{verified-continuation-returns}

RV-PI improves the BeamBC-initialized macro policy using solver-executed returns. Training states are collected from a mixture of the current RV-PI policy, BeamBC, a myopic/regular schedule, and RandomMacro. For each stored state and each feasible action \(q\), the solver is reconstructed at the stored physical state, the candidate \(q\) is applied, and the current macro policy is used for all later decisions until the terminal transition. The return is the negative normalized future squared error:

The training-state distribution is intentionally broader than on-policy data alone. At each improvement cycle, states are collected from the current RV-PI policy, the BeamBC initializer, the regular/myopic schedule, and RandomMacro. The goal is not offline policy evaluation in the formal sense; it is to expose the update to solver states produced by several plausible schedules so that one policy cycle does not overfit to a narrow prefix distribution. For shallow water, six training cases are sampled per source in each cycle; Brusselator uses twelve per source. The physical state tensors and the deployable macro observation are stored so candidate actions can later be re-evaluated from the same state.

For a sampled state, every feasible \(q\) is applied to a newly reconstructed frozen proposal bundle. After that first action, continuation decisions are taken by the current macro policy until the final physical transition. Spatial choices during the entire candidate rollout are always produced by the learned selector. Thus the only oracle quantity in \(G(s_t,q)\) is the error used to score the resulting trajectory; the transition path itself is deployable. This matters because an oracle spatial selector would make the temporal policy appear stronger than the system that can actually be executed.

\begin{equation}
G(s_t,q)=-\frac{\sum_{k=t}^{T-1}\|u_{k+1}^{(q)}-u^\ast_{k+1}\|_2^2}{\sum_{k=t}^{T-1}\|u^\ast_{k+1}\|_2^2}.
\end{equation}

No learned critic determines these target values: candidate actions are evaluated by the actual neural-operator environment. This makes the return expensive during training but removes one important source of policy-evaluation bias.

\subsection{Advantage-weighted conservative update}\label{advantage-weighted-conservative-update}

For the action \(q_{\mathrm{old}}\) taken in the sampled trajectory, verified advantages are defined by

\begin{equation}
A(s_t,q)=G(s_t,q)-G(s_t,q_{\mathrm{old}}).
\end{equation}

Feasible action advantages are converted into a normalized exponential target distribution. The temperature is lower-bounded by a configured value and by the empirical standard deviation of the finite advantages. Alternative actions whose advantage does not exceed a small data-dependent margin are strongly downweighted. The policy is then optimized against this soft target while retaining two conservative terms: cross-entropy to the previous action and KL regularization to the previous policy distribution. In compact notation,

The target construction follows the implementation closely. Let \(\sigma_A\) denote the standard deviation of the finite feasible advantages and \(\tau_0\) the configured minimum temperature. The effective temperature is \(\tau=\max(\tau_0,\sigma_A,10^{-5})\), and raw weights are proportional to \(\exp(\operatorname{clip}(A/\tau,-8,8))\). A data-dependent margin \(m=\max(10^{-4},0.1\sigma_A)\) suppresses alternative actions whose improvement over the behavior action does not exceed \(m\) by multiplying their weight by \(10^{-3}\) before normalization. This small dead zone reduces policy churn caused by nearly indistinguishable verified returns.

\begin{equation}
w_q\propto \exp\!\left(\operatorname{clip}(A(s_t,q)/\tau,-8,8)\right),\qquad \tau=\max(\tau_0,\operatorname{std}(A_{\mathrm{feas}}),10^{-5}),
\end{equation}

followed by the margin suppression described above and normalization over feasible actions. The first term in the policy loss is therefore a cross-entropy to a soft verified-action distribution rather than a hard argmax label. The behavior-cloning term retains the action that generated the state, and the KL term anchors the full distribution to the pre-update policy. The combination is a pragmatic conservative update: it uses measured trajectory advantage aggressively enough to move the policy while reducing the probability that one noisy state batch rewrites the schedule globally.

\begin{equation}
\mathcal{L}=\mathrm{CE}(\pi_{\theta},w(A))+\lambda_{\mathrm{BC}}\mathrm{CE}(\pi_{\theta},q_{\mathrm{old}})+\lambda_{\mathrm{KL}}\mathrm{KL}(\pi_{\mathrm{old}}\|\pi_{\theta}).
\end{equation}

For shallow water, \(\lambda_{\mathrm{BC}}\) = 0.25 and \(\lambda_{\mathrm{KL}}\) = 0.1; the same implementation pattern is used for Brusselator. The update therefore favors actions with measured long-horizon advantage without allowing a single improvement cycle to arbitrarily rewrite the policy.

\subsection{Validation accept/rollback}\label{validation-acceptrollback}

After each policy-improvement cycle, the pre-update and post-update policies are evaluated on the same held-out validation trajectories with identical solver components. The update is accepted only if mean trajectory relative \(L_2\) decreases; otherwise, the old policy parameters are restored. This validation gate is part of the algorithmic design and is applied before final test evaluation. The held-out test set is not used to select improvement cycles.

The acceptance criterion is deliberately external to the minibatch objective. A candidate policy may fit the verified targets yet still interact poorly with the closed-loop solver after its decisions alter the state distribution. We therefore evaluate the old and new policies by complete validation rollouts and compare mean trajectory relative \(L_2\). Only a strict reduction is accepted. Shallow-water RV-PI runs for five candidate cycles, whereas ImmediateOnlyPI runs for three; rejected cycles restore the previous policy parameters. The final test evaluation is performed only with the validation-selected checkpoint.

\subsection{Immediate-only ablation}\label{immediate-only-ablation}

ImmediateOnlyPI retains the same solver components, policy architecture, initialization strategy, state sources, feasible actions, advantage-weighted target construction, regularization, and validation accept/rollback protocol, while evaluating candidate actions using an immediate rather than full-horizon return. Thus,

\begin{equation}
G_{\mathrm{immediate}}(s_t,q)=-\frac{\|u_{t+1}^{(q)}-u^\ast_{t+1}\|_2^2}{\|u^\ast_{t+1}\|_2^2}.
\end{equation}

The RV-PI versus ImmediateOnlyPI comparison therefore directly tests the hypothesis that downstream trajectory consequences contain useful scheduling information beyond immediate correction quality.

\subsection{Why verified rollouts instead of a learned critic}\label{why-verified-rollouts-instead-of-a-learned-critic}

A natural alternative is to train a critic $Q_{\omega}(s,q)$ and use it to rank refinement counts. RV-PI instead exploits the small four-action macro space and the availability of the learned solver offline to evaluate candidate actions by explicit continuation rollout. This changes the statistical problem: rather than relying on a value approximator to extrapolate long-horizon consequences from limited transitions, the method spends additional offline optimization effort to obtain target returns directly from the solver. RV-PI is therefore best understood as approximate policy iteration with rollout verification rather than as a value-function approximation algorithm.

The trade-off is clear. Verified returns scale with the number of sampled states, feasible actions, and remaining horizon, so training can be expensive. At deployment, however, that verification machinery disappears; only the compact macro policy is evaluated. For the present study this asymmetry is attractive because the scientific question concerns whether long-horizon information is useful for allocation. Using solver-executed returns makes that question easier to interpret than using a learned critic whose own approximation error would become an additional explanatory variable.

\subsection*{Algorithm 2. Rollout-verified policy improvement (RV-PI)}
\noindent\textbf{Input:} BeamBC-initialized policy \(\pi\), fixed solver components, training and validation splits.\\
\textbf{For cycle \(=1,\ldots,K\):}
\begin{enumerate}
\item Collect deployable physical states from a mixture of current and baseline policies.
\item For every state, enumerate all feasible \(q\in\{0,1,2,4\}\).
\item Execute each candidate in the learned PDE solver and continue with current \(\pi\) to terminal time.
\item Compute verified long-horizon returns and advantages relative to the action in the collected trace.
\item Convert advantages to conservative soft action targets; update \(\pi\) with BC and KL regularization.
\item Evaluate old and new \(\pi\) on the validation split.
\item Accept the new policy only if validation trajectory Relative \(L_2\) decreases; otherwise rollback.
\end{enumerate}
\noindent\textbf{Return:} the validation-selected macro policy.

\section{Experimental setup}\label{experimental-setup}

\subsection{Governing systems, released benchmark interfaces, and splits}\label{governing-systems-released-benchmark-interfaces-and-splits}

Shallow water. We use the official shallow-water archives distributed with the Laplace Neural Operator benchmark \cite{ref4}. The adapter preserves the 72-frame \(256\times 256\) physical trajectories and the official operator input field \(a(x,y)\). The experiment uses 230 training cases, 20 validation cases, and 50 held-out test cases. The solver advances every fourth native frame, giving 17 learned transitions after the initial state. The budget is \(B=32\) local calls over the trajectory.

Brusselator. We use the forcing-driven Brusselator archive released with the same benchmark family \cite{ref4,ref16}. The state consumed by our adapter is a one-channel \(28\times 28\) field with 39 physical frames; the exposed exogenous input to the transition model is a scalar forcing value at each time. The released training archive supplies 800 training cases. Because the release does not provide a separate validation archive, we partition its 200-case evaluation archive once, using fixed seed 2026, into 100 cases for model/policy selection and 100 cases reserved for final held-out evaluation. The solver predicts all 38 transitions and uses \(B=76\) committed local corrections. Reported Brusselator values therefore correspond to this fixed held-out subset rather than to the entire released evaluation archive.

Governing systems and released interfaces. The shallow-water benchmark is generated from the viscous rotating shallow-water equations on a spherical domain \cite{ref4,ref17}, which may be written in material-derivative form as

\begin{equation}
\frac{D\mathbf{V}}{Dt}=-f\,\mathbf{k}\times\mathbf{V}-g\nabla h+\nu\nabla^2\mathbf{V},
\end{equation}

\begin{equation}
\frac{Dh}{Dt}=-h\nabla\cdot\mathbf{V}+\nu\nabla^2h.
\end{equation}

Here \(\mathbf{V}\) is the tangential velocity field and \(h\) is the fluid-layer height. The released archive is consumed exactly as provided: our adapter denotes the benchmark input array generically as the condition field \(a(x,y)\) and predicts the released trajectory channels without reconstructing unexposed physical variables or imposing additional equations at inference.

The Brusselator benchmark is generated from the two-species reaction-diffusion system \cite{ref4,ref16}

\begin{equation}
\frac{\partial u}{\partial t}=D_0\nabla^2u+a+f(t)-(1+b)u+vu^2,
\end{equation}

\begin{equation}
\frac{\partial v}{\partial t}=D_1\nabla^2v+bu-vu^2.
\end{equation}

The learning interface used in this study follows the released arrays rather than the full latent chemical state: it exposes the scalar forcing together with a single observed field trajectory on the \(28\times 28\) grid. We therefore train and evaluate the neural solver on that released interface and do not reconstruct the unexposed companion species.

\subsection{Training and model-selection protocol}\label{training-and-model-selection-protocol}

All normalization statistics are computed from training data only. On shallow water, the coarse operator uses width 32, 12 Fourier modes, and depth 4; the local operator uses the same width/modes/depth. The field is partitioned into an \(8\times 8\) grid of 32-cell cores with an 8-cell halo. Coarse training uses 12 epochs and AdamW with learning rate \(5\times 10^{-4}\). The local corrector is trained for 6 epochs, and the selector is trained for 16 epochs on labels constructed from 96 training cases. The macro teacher uses 32 training cases with beam width 16. Both BeamBC and the RV-PI policies use learning rate \(5\times 10^{-4}\); RV-PI performs up to five improvement cycles, while ImmediateOnlyPI performs three.

On Brusselator V2, the global FNO uses width 32, 10 modes, and depth 4; the local FNO uses width 32, 8 modes, and depth 3. The \(28\times 28\) field is partitioned into a \(7\times 7\) patch grid with 4-cell cores and a one-cell halo. Coarse training uses the six-step closed-loop objective described in Section 4.2 for 16 epochs. The local corrector is trained for 8 epochs. Selector labels are generated from 320 training cases and the selector is trained for 18 epochs. The macro teacher uses 80 training cases with the same beam width 16. The action space and policy architecture remain unchanged; only the benchmark-specific budget and observation dimensions implied by the field tensors differ.

Checkpoint selection is always based on validation data. The coarse Brusselator checkpoint is selected by validation trajectory relative \(L_2\) on a fixed 32-case subset during training. Policy cycles are selected by full validation rollouts. No test metric is used to choose a horizon, cycle, checkpoint, or random seed. The shallow-water backbone and deterministic baselines are frozen once trained and then reused across policy seeds; similarly, Brusselator V2 policies share the same stabilized global/local/selector components. This is why some rows in Table 2 have zero standard deviation across policy seeds.

\begin{table}[!htbp]
\centering
\caption{Experimental and solver configuration.}
\label{tab:config}
\small
\begin{tabularx}{\textwidth}{>{\raggedright\arraybackslash}p{0.24\textwidth}XX}
\toprule
\textbf{Setting} & \textbf{Shallow water} & \textbf{Brusselator V2} \\
\midrule
Native trajectory & 72 frames, \(256\times 256\) & 39 frames, \(28\times 28\) \\
Train / val / test & 230 / 20 / 50 & 800 / 100 / 100 \\
Learned transitions & 17 (stride 4) & 38 \\
Global FNO & width 32, modes 12, depth 4 & width 32, modes 10, depth 4 \\
Local FNO & width 32, modes 12, depth 4 & width 32, modes 8, depth 3 \\
Patch partition & \(8\times 8\); core 32; halo 8 & \(7\times 7\); core 4; halo 1 \\
Selector & Transformer, width 64, 4 heads, 2 layers & Transformer, width 64, 4 heads, 2 layers \\
Macro action space & \{0,1,2,4\} & \{0,1,2,4\} \\
Trajectory budget & 32 local calls & 76 local calls \\
Policy seeds & 42, 123, 2026 & 42, 123, 2026 \\
Coarse training & 12 epochs; 1-step + 2-step scheduled closed-loop term & 16 epochs; K=6 full closed-loop + one-step anchor \\
Selector labels / epochs & 96 train cases / 16 epochs & 320 train cases / 18 epochs \\
Beam teacher & 32 train cases; width 16 & 80 train cases; width 16 \\
Policy improvement & RV-PI 5 cycles; Immediate 3 & RV-PI 5 cycles; Immediate 3 \\
\bottomrule
\end{tabularx}
\end{table}

\subsection{Baselines and fairness}\label{baselines-and-fairness}

We compare the complete RV-PI policy with controls chosen to answer distinct questions. CoarseOnly uses no applied local correction. RandomMacro samples feasible refinement counts and tests whether arbitrary temporal placement is already sufficient. UniformMacro uses the regular action closest to q=2 subject to exact remaining-budget feasibility. MaxFeasibleMacro repeatedly chooses the largest currently feasible action and serves as an aggressive early-spending heuristic. SetAwareMyopicMacro retains the same learned spatial selector but uses a short-sighted deployable temporal rule; it is included as a secondary scheduling control rather than as the principal horizon ablation. BeamBC is the behavior-cloned train-only beam-search policy. ImmediateOnlyPI is the controlled policy-improvement ablation described above. All adaptive methods within a benchmark share the same trained global operator, local corrector, selector, frozen-proposal semantics, action set, and total applied-refinement budget. The central comparison with ImmediateOnlyPI therefore targets temporal credit assignment rather than the correction mechanism itself.

\subsection{Metrics and reporting protocol}\label{metrics-and-reporting-protocol}

The primary metric is full-trajectory relative \(L_2\), computed by concatenating predicted and ground-truth states over the physical rollout and taking the global vector-norm ratio. We also report final-frame relative \(L_2\) and the number of local calls. Policy-training variability is measured with seeds 42, 123, and 2026. Operator and deterministic baseline checkpoints are frozen; therefore zero standard deviation for these rows reflects reuse of the same checkpoint/schedule rather than three independent operator retrainings. Each validation-selected policy is evaluated once on the fixed held-out test split for its seed.

\subsection{Reproducibility and seed scope}\label{reproducibility-and-seed-scope}

Seeds 42, 123, and 2026 index stochastic policy training and RandomMacro schedules; the global/local operators and deterministic baselines are frozen once per benchmark. Reported policy variability therefore reflects temporal-policy optimization on a common solver backbone. Each validation-selected policy is evaluated once on the fixed held-out test split, while matched adaptive methods use the same trained solver components and exact refinement-intervention budget.

The repository records seed-level comparison tables, macro action timelines, selector diagnostics, and Brusselator stability metrics. Reported experiments hold solver components fixed and compare temporal allocation under matched refinement-intervention budgets.

\section{Results}\label{results}

\subsection{Deployment-aligned training establishes a stable autoregressive backbone}\label{deployment-aligned-training-establishes-a-stable-autoregressive-backbone}

Brusselator provides a direct test of whether the learned transition model is suitable for closed-loop allocation. With one-step-oriented training, validation one-step relative \(L_2\) is 0.3412 while the 38-step trajectory relative \(L_2\) reaches 8.2505 and the final-frame error reaches 84.71. The V2 multi-step protocol reduces one-step relative \(L_2\) to 0.3163 and, more importantly, reduces trajectory relative \(L_2\) to 0.09369 and final-frame relative \(L_2\) to 0.1082. Exposure to self-generated states therefore converts the coarse operator into a stable autoregressive backbone without trading away one-step fidelity.

Figure 2 makes this separation visible. The temporal behavior changes from rapidly growing recursive error to a bounded rollout, while the one-step metric remains in the same accuracy regime. This solver-quality gate matters because the refinement policy is intended to optimize how a functioning learned solver spends corrective interventions. All subsequent Brusselator comparisons therefore use the same stabilized V2 backbone, so differences among policies reflect allocation rather than changes to the transition model.

Relative to the one-step-oriented model, V2 reduces validation trajectory error by 98.86\% and final-frame error by 99.87\%, while improving one-step relative \(L_2\) by 7.30\%. This result is consistent with the importance of deployment-aligned closed-loop training: the substantial gain is obtained without increasing operator width or depth and accompanies direct exposure to self-generated states during multi-step optimization \cite{ref18,ref19,ref20}.

\begin{figure}[!t]
\centering
\includegraphics[width=\textwidth]{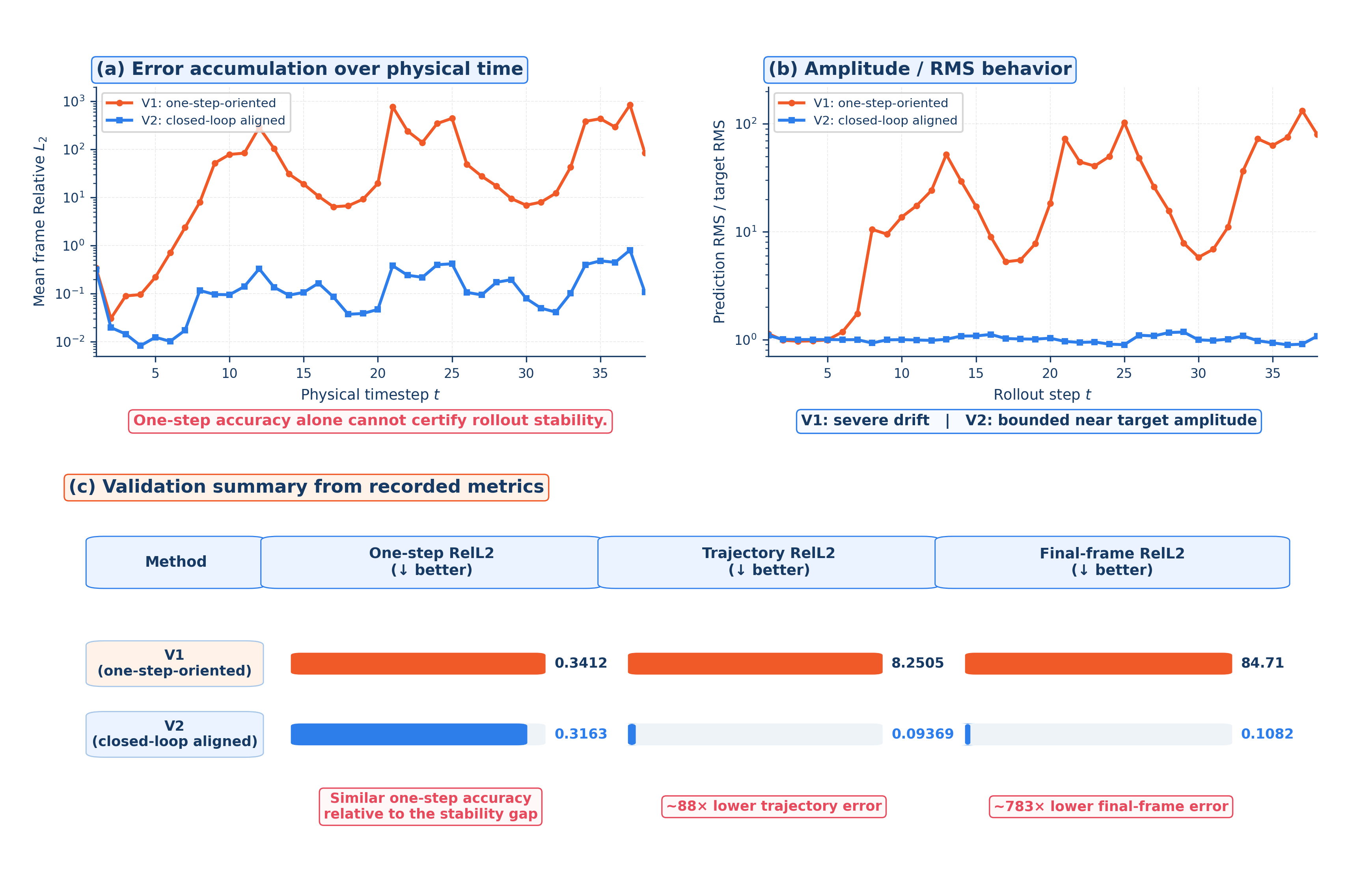}
\caption{Recorded Brusselator validation behavior under one-step-oriented (V1) and deployment-aligned closed-loop (V2) operator training. Panels (a)-(b) plot the measured step-wise mean Relative \(L_2\) and prediction-to-target RMS ratio over the 38-transition rollout on logarithmic axes; panel (c) reports the corresponding validation summary metrics. The figure is generated directly from the recorded diagnostic tables.}
\label{fig:stability}
\end{figure}

\subsection{Main fixed-budget results}\label{main-fixed-budget-results}

Table 2 and Figure 3 summarize the central result. On shallow water, RV-PI achieves 0.6910 +/- 0.0088 mean trajectory relative \(L_2\). The strongest stochastic non-RL baseline, RandomMacro, obtains 0.7081 +/- 0.0054, corresponding to a 2.41\% mean error reduction for RV-PI. The controlled long-horizon comparison is larger: ImmediateOnlyPI obtains 0.73025 +/- 0.00057, so full-horizon RV-PI lowers error by 5.37\%. RV-PI also improves over BeamBC by 6.54\% and over CoarseOnly by approximately 6.66\%. The final-frame metric follows the same direction on this benchmark: RV-PI reaches 0.9821 +/- 0.0358, the lowest among the reported methods.

The ranking on shallow water is especially informative because RandomMacro is already strong. Relative to CoarseOnly, random scheduling lowers mean trajectory error from 0.74035 to 0.70809. RV-PI lowers it further to 0.69103 while using the same 32 committed corrections. The 2.41\% margin over RandomMacro is smaller than the 5.37\% margin over ImmediateOnlyPI, but it is scientifically valuable precisely because the random baseline is not trivial: it shows that the local corrector itself contributes substantial value and that RV-PI extracts an additional gain from temporal placement.

The Brusselator regime is different. CoarseOnly is already accurate at 0.10165, and several adaptive schedules can worsen the result if corrections are placed poorly. RandomMacro reaches 0.10513 and the aggressive deterministic schedules are worse still. Against that background, RV-PI reaches 0.09954. The improvement is numerically smaller in absolute terms than on shallow water but qualitatively stronger as an allocation result: refinement is not automatically beneficial in this low-error regime, yet a future-aware policy can make the same correction mechanism improve the trajectory on average.

On Brusselator V2, RV-PI reaches a three-seed mean trajectory relative \(L_2\) of 0.099542 +/- 0.001997, the lowest value in the comparison. ImmediateOnlyPI obtains 0.101896 +/- 0.001722, so the long-horizon return yields a 2.31\% mean reduction. RV-PI also improves over BeamBC by 1.72\%, CoarseOnly by 2.07\%, RandomMacro by 5.32\%, and SetAwareMyopicMacro by 9.21\%. The gain is smaller than on shallow water against ImmediateOnlyPI, but its sign is preserved across a substantially different dynamical system and a different local-refinement geometry.

\begin{table}[!htbp]
\centering
\caption{Three-seed fixed-budget comparison. Lower trajectory and final-frame Relative \(L_2\) are better. 'Local calls' denotes the number of committed patch corrections; deterministic rows reuse the same frozen solver/checkpoint across policy seeds, so zero standard deviation does not imply independent backbone retraining.}
\label{tab:results}
\scriptsize
\setlength{\tabcolsep}{3.2pt}
\resizebox{\textwidth}{!}{%
\begin{tabular}{llrrrrr}
\toprule
\textbf{PDE} & \textbf{Method} & \textbf{Trajectory RelL2 mean} & \textbf{Trajectory std} & \textbf{Final RelL2 mean} & \textbf{Final std} & \textbf{Local calls} \\
\midrule
Shallow water & CoarseOnly & 0.740347 & 0.000000 & 1.056560 & 0.000000 & 0 \\
Shallow water & Random\allowbreak Macro & 0.708093 & 0.005420 & 1.020698 & 0.011187 & 32 \\
Shallow water & BeamBC & 0.739403 & 0.000000 & 1.080303 & 0.000000 & 32 \\
Shallow water & ImmediateOnly\allowbreak PI & 0.730245 & 0.000571 & 1.082184 & 0.000879 & 32 \\
Shallow water & SetAwareMyopic\allowbreak Macro & 0.737156 & 0.000000 & 1.017257 & 0.000000 & 32 \\
Shallow water & RV-PI & 0.691026 & 0.008779 & 0.982092 & 0.035847 & 32 \\
Brusselator V2 & CoarseOnly & 0.101647 & 0.000000 & 0.134663 & 0.000000 & 0 \\
Brusselator V2 & Random\allowbreak Macro & 0.105131 & 0.000832 & 0.156753 & 0.023193 & 76 \\
Brusselator V2 & BeamBC & 0.101283 & 0.000000 & 0.161804 & 0.000000 & 76 \\
Brusselator V2 & ImmediateOnly\allowbreak PI & 0.101896 & 0.001722 & 0.140732 & 0.018659 & 76 \\
Brusselator V2 & SetAwareMyopic\allowbreak Macro & 0.109639 & 0.000000 & 0.232940 & 0.000000 & 76 \\
Brusselator V2 & RV-PI & 0.099542 & 0.001997 & 0.155273 & 0.022721 & 76 \\
\bottomrule
\end{tabular}}
\end{table}

\begin{figure}[!htbp]
\centering
\includegraphics[width=0.98\textwidth]{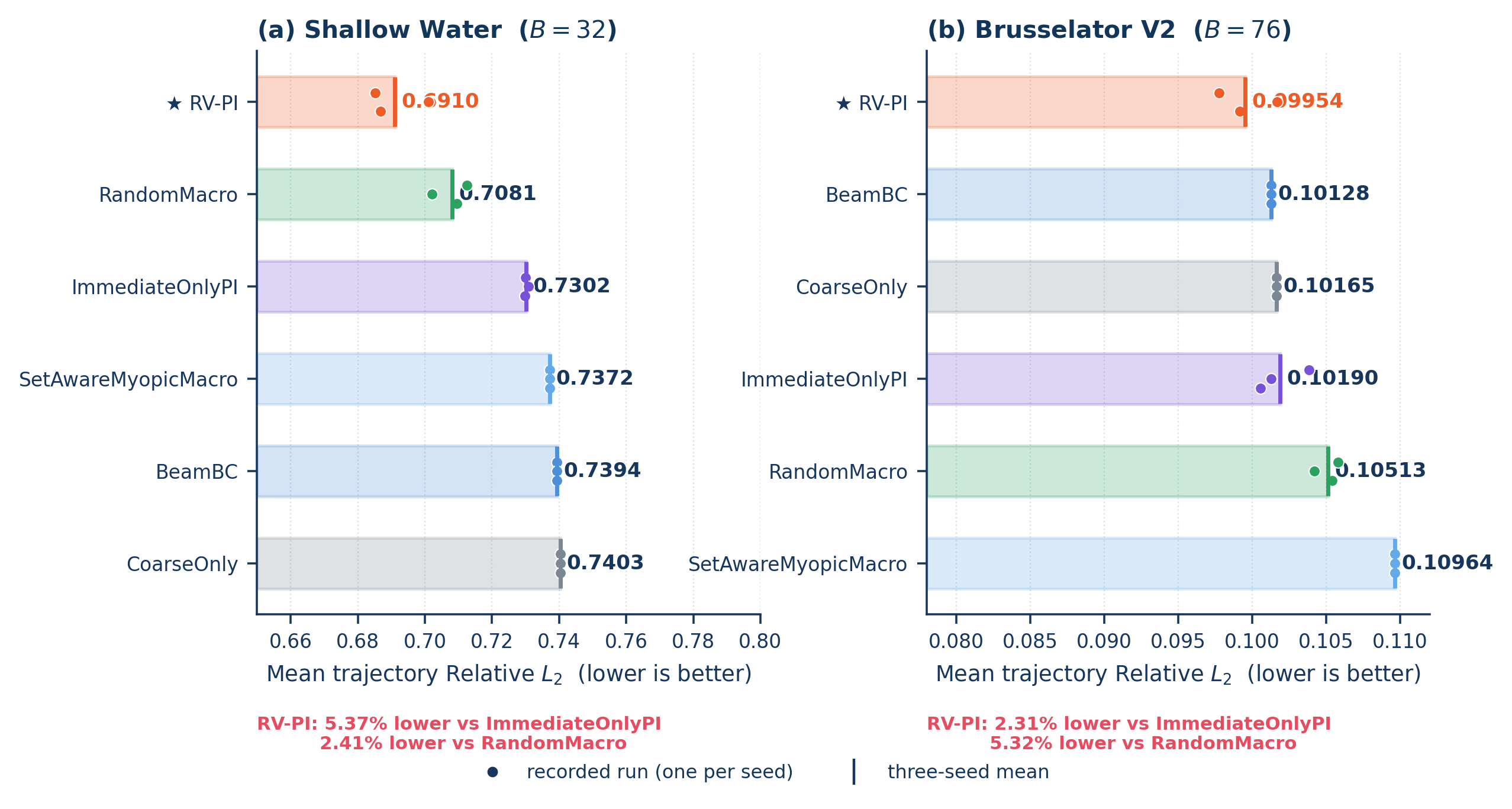}
\caption{Cross-PDE trajectory Relative \(L_2\) under matched refinement-intervention budgets. Bars show three-seed means and markers show the recorded seed-level values. RV-PI has the lowest mean trajectory error on both benchmarks. Within each benchmark, all adaptive methods commit the same number of local corrections.}
\label{fig:results}
\end{figure}

\subsection{The contribution comes from temporal allocation, not extra applied refinements}\label{the-contribution-comes-from-temporal-allocation-not-extra-applied-refinements}

The applied-refinement counts make the allocation comparison explicit. Every adaptive shallow-water method commits 32 local corrections and every adaptive Brusselator method commits 76. The gains in Table 2 therefore arise from temporal placement of the same number of committed corrections. RandomMacro is particularly informative on shallow water: random scheduling already improves substantially over CoarseOnly, so RV-PI's additional 2.41\% reduction demonstrates value from structured long-horizon scheduling on top of a useful correction mechanism.

\subsection{Full-horizon returns outperform immediate returns on average}\label{full-horizon-returns-outperform-immediate-returns-on-average}

The ImmediateOnlyPI ablation is the most direct test of the paper's hypothesis because it preserves the same BeamBC initialization, spatial selector, action set, budget semantics, policy architecture, and validation gate while truncating candidate value to the current transition. On shallow water, RV-PI reduces the three-seed mean trajectory error from 0.73025 to 0.69103, a 5.37\% improvement. On Brusselator V2, the corresponding mean decreases from 0.101896 to 0.099542, a 2.31\% improvement, with two of the three individual policy seeds also improving. The same controlled change - extending credit assignment over the future solver trajectory - therefore benefits mean trajectory accuracy on both benchmark families.

This controlled ablation is more diagnostic than comparison with CoarseOnly alone. The local correction mechanism, spatial selector, and available intervention count are already fixed before the temporal policy is compared. RV-PI changes only how candidate q actions are valued over time. Its advantage over ImmediateOnlyPI therefore provides direct evidence that downstream consequences of a correction contain useful information beyond immediate error reduction.

The Brusselator result is especially informative because the stabilized CoarseOnly trajectory error is already low (0.10165) and several fixed or heuristic schedules degrade it. In this regime, more intervention is not automatically better; the benefit depends on when the fixed correction budget is committed. RV-PI is the only reported method to reduce the three-seed mean below 0.10, reinforcing the interpretation that the policy is learning selective temporal allocation rather than simply activating the local corrector more often.

\subsection{Cross-PDE interpretation of the allocation result}\label{cross-pde-interpretation-of-the-allocation-result}

The value of the second benchmark is not that shallow water and Brusselator cover all relevant PDE regimes. It is that the same allocation mechanism is tested under two substantially different solver geometries. Shallow water uses a \(256\times 256\) native field, an \(8\times 8\) patch partition, 17 learned transitions, and a static released condition field. Brusselator uses a \(28\times 28\) field, a \(7\times 7\) patch partition, 38 transitions, and time-varying forcing. The budgets differ by more than a factor of two and the baseline error scales are very different. Despite these changes, the mean ordering \(\mathrm{RV\text{-}PI}<\mathrm{ImmediateOnlyPI}\) is preserved.

This cross-PDE agreement supports a focused solver-level principle: when local corrections alter the state that will be propagated autoregressively, temporal credit assignment beyond the next frame can improve how a fixed refinement budget is used. The effect persists across changes in patch count, horizon length, condition representation, native spatial resolution, and baseline error scale, which is precisely the transfer axis targeted by the formulation.

The mechanism is consistent with the sequential formulation. ImmediateOnlyPI treats a local correction as valuable only through the next frame. RV-PI can favor an action with modest immediate effect if it moves the solver into a state from which later global predictions are easier to correct, or conserve budget at a locally difficult frame if spending the same calls later produces a larger reduction in accumulated trajectory error. The learned policy therefore optimizes the path of the autoregressive solver rather than a collection of independent one-step predictions.

\subsection{Where the local corrections are applied}\label{where-the-local-corrections-are-applied}

Spatial selection is deliberately separated from macro scheduling. The selector is conditioned on the already selected set, so later patch choices can account for redundancy and overlap rather than reproduce an independent top-k ranking. Figure 4 shows this decomposition using a recorded held-out shallow-water trajectory (case 250, seed 42). Panel (a) compares the actual ImmediateOnlyPI and RV-PI temporal schedules under the same 32-correction budget. Panel (b) shows the corresponding frame-error evolution together with the RV-PI patch set selected at transition 13, where \(q_t=4\), the cumulative patch-use frequency, and the most-used patches on the same trajectory.

The recorded behavior makes the hierarchy concrete. ImmediateOnlyPI spends correction budget from the beginning of the trajectory and exhausts it by transition 12, whereas RV-PI preserves substantial budget for transitions 13-16. At transition 13, RV-PI selects four frozen proposals (patches 46, 14, 62, and 5 in the repository indexing), and the trajectory-level frequency map shows repeated use of a compact subset of spatial locations. The macro policy therefore changes WHEN/HOW MUCH computation is spent, while the set-aware selector resolves WHERE those corrections are committed.

The qualitative trace also matches the quantitative result: after the two schedules diverge strongly in the second half of the rollout, RV-PI maintains lower late-stage frame error on the first held-out shallow-water case (case 250) while using exactly the same total number of corrections. Figure 4 is not a separate benchmark; it is a recorded mechanism view of the same inference procedure summarized numerically in Table 2 and Figure 3.

\begin{figure}[!t]
\centering
\includegraphics[width=\textwidth]{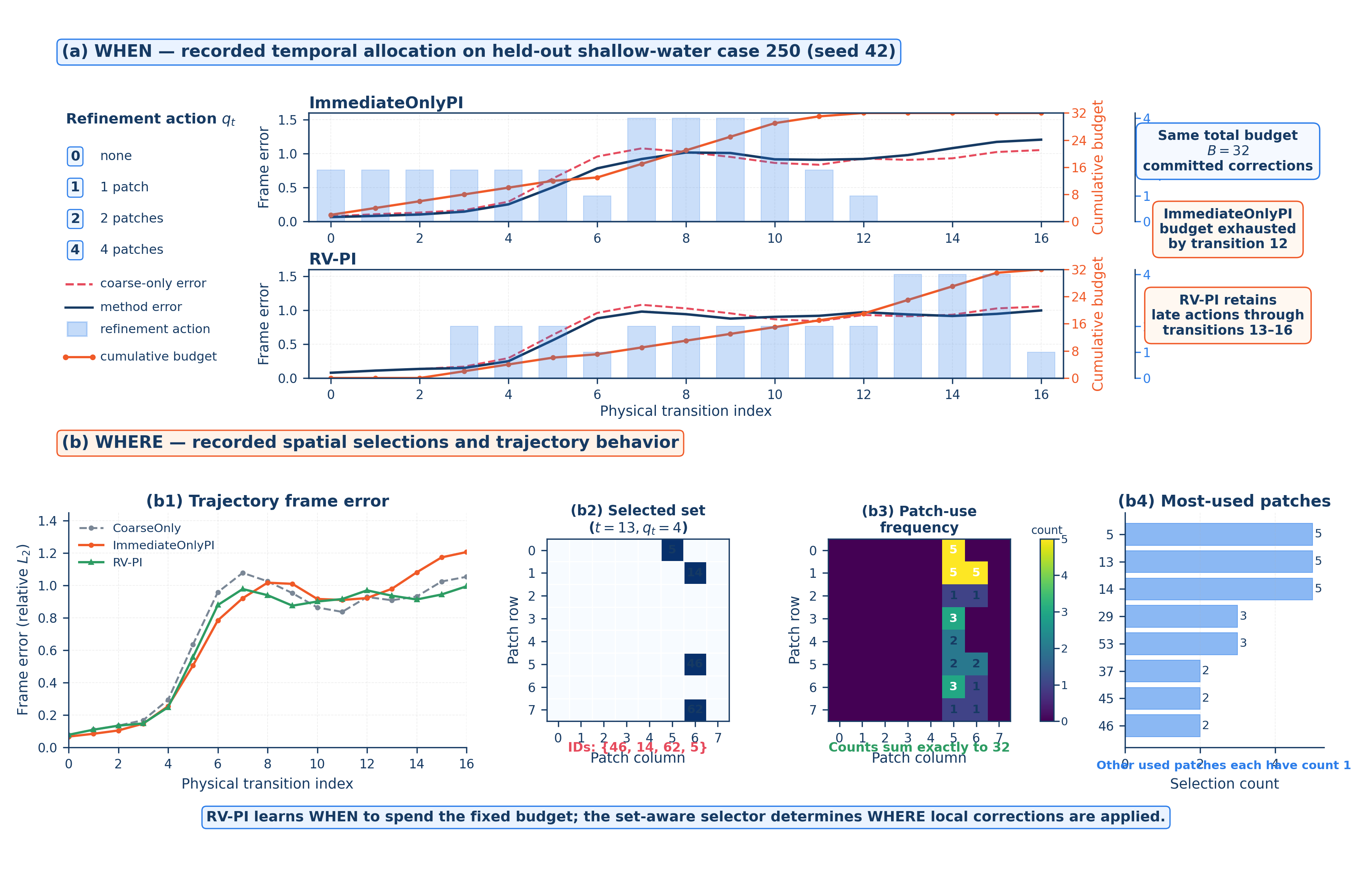}
\caption{Recorded WHEN/WHERE behavior on the first held-out shallow-water case (case 250, seed 42). ImmediateOnlyPI and RV-PI both commit 32 patch corrections but distribute them differently over the 17 learned transitions. Panel (a) shows temporal allocation and cumulative budget use. Panel (b) shows trajectory error, the four-patch set selected by RV-PI at transition 13, cumulative spatial selection frequency, and the most-used patches.}
\label{fig:behavior}
\end{figure}

\section{Discussion}\label{discussion}

\subsection{Solver-level novelty: long-horizon temporal refinement allocation}\label{solver-level-novelty-long-horizon-temporal-refinement-allocation}

The strongest interpretation of the results is solver-level. The global FNO, local FNO, conditional spatial selector, macro action space, and total committed-correction budget are held fixed in the central comparisons. The observed gain is therefore associated with temporal placement of the available correction interventions according to measured future consequences. This shifts the focus from representational capacity alone to a complementary question: how should a learned solver schedule a fixed number of corrective interventions over the physical trajectory?

The central novelty is the long-horizon temporal allocation procedure at solver level. Frozen candidate semantics, simultaneous normalized blending, exact budget feasibility, and conditional spatial selection provide the well-defined deployment substrate needed to make each macro action \(q_t\) comparable. RV-PI then evaluates those macro choices through solver-executed future rollouts. The ImmediateOnlyPI comparison targets the value of extending temporal credit assignment beyond the current transition, while the spatial selector is treated as an enabling component rather than as a separately validated standalone contribution.

\subsection{Relation to adaptive mesh refinement}\label{relation-to-adaptive-mesh-refinement}

The conceptual connection to RL-based AMR is the recognition that refinement decisions are sequential \cite{ref8,ref9}. The computational object being refined is different. AMR changes numerical resolution or discretization. Our solver leaves the physical grid unchanged and applies learned residual corrections on selected fixed-grid patches. This makes the method applicable when the inference engine is a neural surrogate rather than a classical discretization and when remeshing is undesirable or unavailable. It also creates a natural hierarchy: the macro policy handles temporal resource allocation, while the set-aware selector handles spatial localization.

The distinction from time-adaptive operator learning is equally important. Methods such as TANTE \cite{ref23} adapt the temporal step itself, which can reduce the number of global transitions in smooth regions and resolve rapidly varying dynamics with smaller steps. RV-PI keeps the benchmark clock fixed and allocates correction intensity over that clock. The two strategies are complementary: adaptive time stepping changes when the global state is advanced, whereas RV-PI changes when additional local corrective capacity is committed.

The distinction from hybrid numerical correction is also one of control granularity. Super-fidelity warm starts \cite{ref22} and neural-operator/relaxation hybrids \cite{ref12} invoke a classical iterative procedure to recover accuracy or convergence. Here the correction mechanism is itself learned and local, and the policy controls which learned residuals are committed. The method is therefore most relevant when a global learned transition and a local learned corrector are already available as deployable modules.

\subsection{Why frozen proposals matter}\label{why-frozen-proposals-matter}

Without frozen proposals, a multi-patch action would be ambiguous. Sequentially applying one correction and then recomputing all remaining local candidates changes both the state and the computational cost of subsequent choices. The macro action \(q_t\) would no longer represent \(q_t\) comparable refinement calls from a common state, and an apparent gain could arise from hidden extra local computation. Freezing all proposals before selection removes this confound. Simultaneous normalized overlap blending similarly avoids an arbitrary dependence on the order in which overlapping patches are written back.

Frozen proposals also simplify credit assignment. If candidate \(p\) were regenerated after applying candidate \(r\), then the action value of selecting \(p\) would implicitly depend on whether \(r\) happened to be chosen earlier. The macro action \(q_t\) would no longer correspond to a stable set size, and selector labels computed from one ordering would not describe another. By freezing the candidate pool, the only combinatorial interaction is through blending and overlap, which the selected-set mask and transformer can represent explicitly.

\subsection{Design lessons from controlled diagnostics}\label{design-lessons-from-controlled-diagnostics}

The final solver architecture is best understood as the result of three controlled diagnostics rather than as an accumulation of independent modules. First, local predictive accuracy is not a sufficient deployment criterion for an autoregressive solver. Figure~\ref{fig:stability} shows that the one-step-oriented Brusselator model remains in a similar one-step error regime yet becomes unstable over the closed-loop trajectory, whereas deployment-aligned multi-step training produces a bounded rollout. The practical lesson is that temporal-allocation experiments are only meaningful after the backbone itself has passed a long-horizon stability check; otherwise, apparent policy effects can be confounded with recursive model failure.

Second, the value of refinement is intrinsically time dependent. ImmediateOnlyPI preserves the same global operator, local corrector, selector, action set, budget, policy architecture, and validation gate as RV-PI, but values a candidate only through the current transition. Its higher mean trajectory error on both benchmarks therefore exposes a specific failure mode of reactive allocation: an intervention can look favorable locally while consuming budget that would have greater downstream value. The recorded shallow-water trace in Figure~\ref{fig:behavior} makes this mechanism visible. ImmediateOnlyPI spends from the beginning and exhausts the 32-correction budget by transition 12, whereas RV-PI retains actions for later transitions. The lesson is not that more policy complexity is automatically beneficial, but that delayed consequences must be measurable when corrections alter states that will themselves be propagated.

Third, long-horizon credit assignment is only interpretable when refinement actions have stable semantics. Freezing local proposals before set construction ensures that an action requesting \(q_t\) corrections always refers to \(q_t\) candidates generated from the same provisional state. Simultaneous normalized blending then prevents the applied correction from depending on an arbitrary write order. These choices are therefore not merely implementation conveniences: they remove hidden changes in candidate definition and computational work that would otherwise contaminate comparisons among macro actions.

Taken together, these diagnostics suggest a practical hierarchy for budgeted learned PDE solvers: first establish a stable closed-loop backbone, then define a deployment-valid correction action with comparable cost and semantics, and only then optimize when the finite correction budget should be spent. This hierarchy also sharpens the scope of the present result: RV-PI addresses temporal allocation once the underlying transition and local-correction mechanisms are already usable, rather than compensating for an unstable surrogate or an ill-defined refinement operator.

\subsection{Primary endpoint: trajectory-level accuracy}\label{primary-endpoint-trajectory-level-accuracy}

The primary endpoint follows the optimization objective used by RV-PI: normalized error accumulated over the propagated trajectory. This metric is natural for repeated-query forecasting and digital-twin-style state tracking because every intermediate state contributes both to application accuracy and to the distribution seen by later transitions. Table 2 also reports final-frame relative \(L_2\) as a secondary descriptive metric, while model selection and the central method comparisons are defined by trajectory-level accuracy.

The objective can be adapted without changing the solver architecture. Applications centered on a terminal state can replace the integrated return with a terminal-weighted objective, whereas interval forecasting can retain the trajectory-integrated criterion used here. This separation between solver architecture and task-level return is one practical benefit of the RV-PI formulation: the temporal policy can be optimized for the quantity of interest while the global operator, local corrector, and spatial selector remain unchanged.

\subsection{Scope and practical implications}\label{scope-and-practical-implications}

The formulation targets fixed-grid neural-operator solvers with a pre-trained local correction mechanism and a hard intervention budget. The context input can follow the benchmark: shallow water uses a static released condition field, whereas Brusselator uses time-varying forcing. The macro policy acts on solver state, selector statistics, and budget context instead of hand-coded PDE-specific error estimators, making the same decision architecture usable across distinct operator inputs and rollout geometries.

\subsection{Evidence for long-horizon refinement allocation}\label{evidence-for-long-horizon-refinement-allocation}

The evidence chain is aligned with the paper's central claim. First, the autoregressive backbone remains stable under closed-loop deployment. Second, frozen local proposals and a ground-truth-free conditional selector provide a deployable spatial correction mechanism. Third, every compared adaptive trajectory satisfies the same exact committed-correction budget. Under those controls, full-horizon RV-PI achieves lower three-seed mean trajectory Relative \(L_2\) than ImmediateOnlyPI on both benchmark families.

The result is strongest as a mechanism statement rather than as a catalogue of benchmark wins. Shallow water shows a 5.37\% mean reduction relative to ImmediateOnlyPI, while the lower-error Brusselator regime shows a 2.31\% reduction. The shared direction under substantially different horizons, patch geometries, and context representations supports the value of future-aware temporal credit assignment in budgeted learned PDE solvers.

\subsection{Practical use and extension paths}\label{practical-use-and-extension-paths}

The framework is most natural in repeated-query settings where offline optimization can be amortized: forecasting ensembles, parameter studies, surrogate-assisted design, or digital-twin-style state propagation. Once trained, the global/local operators and selector can be reused while the macro policy adds only a small decision network. The policy itself is also modular: changing the application objective from integrated field error to a weighted quantity of interest would require changing the verified return and validation metric, not redesigning the solver architecture.

Two implementation extensions follow naturally from the current semantics. Lazy or cached proposal generation can reduce the cost of materializing unused candidates, while richer policy observations can incorporate deployable uncertainty or physics-residual diagnostics when available. Both preserve the core WHERE/WHEN decomposition and exact-budget semantics established here.

RV-PI moves the expensive verification step to offline policy optimization. At inference, the macro network adds only a compact convolutional/MLP decision pass, while the committed local-correction budget remains identical to the matched baselines. This makes the approach particularly suitable for repeated-query settings in which offline solver and policy training can be amortized over many deployments.

\section{Conclusion}\label{conclusion}

We introduced a budgeted adaptive neural-operator solver that separates spatial refinement from temporal allocation while preserving a single closed-loop PDE trajectory. A global FNO advances the full field, a local FNO produces a frozen pool of patch residuals, a conditional transformer selector constructs the spatial patch set, and a compact macro policy decides how many corrections to commit at each physical transition. Exact feasibility masking preserves the committed-correction budget, and simultaneous normalized blending gives multi-patch actions order-independent semantics. This organization isolates a solver-level question that operator architecture alone does not answer: how should a fixed refinement-intervention budget be distributed over physical time?

The policy is improved through solver-executed continuation returns rather than a learned critic. On the official shallow-water benchmark, full-horizon RV-PI reduces three-seed mean trajectory relative \(L_2\) from 0.73025 for ImmediateOnlyPI to 0.69103, a 5.37\% reduction, and also improves on RandomMacro by 2.41\%. On the stabilized Brusselator V2 benchmark, the corresponding mean changes from 0.10190 to 0.09954, a 2.31\% reduction, while RandomMacro is improved by 5.32\%. Deployment-aligned multi-step training additionally reduces Brusselator validation trajectory relative \(L_2\) from 8.2505 to 0.09369, establishing the stable autoregressive environment required for meaningful allocation.

Across the evaluated fixed-grid settings, the results establish a consistent design principle: adaptive learned PDE solving is not only a question of WHERE local correction should be applied, but also WHEN committed local corrections are most valuable along the propagated trajectory. By separating conditional spatial selection from long-horizon temporal credit assignment, the proposed solver turns a fixed refinement-intervention budget into a sequential optimization problem whose objective matches the trajectory used at deployment.

\section*{Data availability}

The shallow-water and Brusselator arrays are the released datasets distributed with the Laplace Neural Operator study \cite{ref4}. The present implementation preserves the benchmark-provided arrays and exposes them through repository adapters. For Brusselator, the released 200-case evaluation archive is partitioned once into a 100-case validation subset and a disjoint 100-case final held-out subset as described in Section 6.1. No final held-out targets are used to train the macro policy or spatial selector.

\section*{Code availability}

The implementation, configurations, result tables, evaluation scripts, and released checkpoints are publicly available in the RVPI-PDE repository at \url{https://github.com/Alfred-Duncan/RVPI-PDE}.

\section*{Declaration of competing interest}

The author declares that there are no known competing financial interests or personal relationships that could have appeared to influence the work reported in this paper.

\section*{CRediT authorship contribution statement}

Ange Tong: Conceptualization, Methodology, Software, Validation, Formal analysis, Investigation, Visualization, Writing - original draft, Writing - review \& editing.

\clearpage\section*{Declaration of generative AI and AI-assisted technologies in the manuscript preparation process}

During the preparation of this work, the author used OpenAI ChatGPT to assist with language editing, organization, and presentation. After using this tool, the author reviewed and edited the content as needed, verified the scientific statements, numerical results, references, and final quantitative figures against the underlying implementation and recorded outputs, and takes full responsibility for the content of the manuscript.

\end{document}